\documentclass[10pt,twocolumn,letterpaper]{article}

\usepackage[pagenumbers]{cvpr} 

\usepackage{amsthm}           
\usepackage{algorithm}        
\usepackage{algorithmic}      
\usepackage{multirow}         
\usepackage{tabularx}         
\usepackage{makecell}         
\usepackage{colortbl}         

\definecolor{GroupBlue}{RGB}{220,240,255}

\definecolor{cvprblue}{rgb}{0.21,0.49,0.74}
\usepackage[pagebackref,breaklinks,colorlinks,allcolors=cvprblue]{hyperref}
\usepackage{bm}

\title{ProFocus: Proactive Perception and Focused Reasoning in \\ Vision-and-Language Navigation}

\author{
    Wei Xue\textsuperscript{1} \quad
    Mingcheng Li\textsuperscript{1} \quad
    Xuecheng Wu\textsuperscript{3} \quad
    Jingqun Tang\textsuperscript{4} \\
    Dingkang Yang\textsuperscript{1,2,$\dagger$,\S} \quad
    Lihua Zhang\textsuperscript{1,2,$\dagger$} \\
    \footnotesize \textsuperscript{$\dagger$}corresponding authors \quad \textsuperscript{\S}project lead \\
    \textsuperscript{1}College of Intelligent Robotics and Advanced Manufacturing, Fudan University \\
    \textsuperscript{2}Fysics Intelligence Technologies Co., Ltd. (Fysics AI) \\
    \textsuperscript{3}Xi'an Jiaotong University \quad
    \textsuperscript{4}ByteDance \\
    {\tt\small wxue24@m.fudan.edu.cn, \{dkyang20, lihuazhang\}@fudan.edu.cn}
}

\begin{document}
\maketitle

\begin{abstract}
Vision-and-Language Navigation (VLN) requires agents to accurately perceive complex visual environments and reason over navigation instructions and histories. However, existing methods passively process redundant visual inputs and treat all historical contexts indiscriminately, resulting in inefficient perception and unfocused reasoning.
To address these challenges, we propose \textbf{ProFocus}, a training-free progressive framework that unifies \underline{Pro}active Perception and \underline{Focus}ed Reasoning through collaboration between large language models (LLMs) and vision-language models (VLMs).
For proactive perception, ProFocus transforms panoramic observations into structured ego-centric semantic maps, enabling the orchestration agent to identify missing visual information needed for reliable decision-making, and to generate targeted visual queries with corresponding focus regions that guide the perception agent to acquire the required observations.
For focused reasoning, we propose Branch-Diverse Monte Carlo Tree Search (BD-MCTS) to identify top-$k$ high-value waypoints from extensive historical candidates. The decision agent focuses reasoning on the historical contexts associated with these waypoints, rather than considering all historical waypoints equally.
Extensive experiments validate the effectiveness of ProFocus, achieving state-of-the-art performance among zero-shot methods on R2R and REVERIE benchmarks.

\end{abstract}    
\section{Introduction}
\label{sec:intro}
Vision-and-Language Navigation (VLN)~\cite{anderson2018vision} enables agents to explore physical environments by following natural language instructions. Traditional training-based approaches~\cite{anderson2018vision,hao2020towards,chen2022think}, while achieving progress, remain data-intensive and environment-dependent~\cite{lin2025evolvenav}. The rapid development of large language models (LLMs)~\cite{niu2025research} and vision-language models (VLMs)~\cite{qin2025survey} has enabled foundation model-based VLN~\cite{chen2025affordances,zhang2024visionandlanguage} through post-training adaptation~\cite{lin2025navcot,yu2025correctnav} or zero-shot prompting~\cite{zhou2024navgpt,chen2024mapgpt,yin2025unigoal}. Despite these advances, current foundation model-based VLN methods~\cite{zhou2024navgpt,
yin2025unigoal,chen2025affordances} encounter two key limitations. (1) \textbf{Passive visual perception.} For VLM-driven research~\cite{10657191,lin2025evolvenav,zhang2025cross,chen2024mapgpt}, agents process panoramic or multi-view visual observations~\cite{yu2025correctnav,zhang2025embodied,song2025towards} whose redundancy inflates visual-token counts and causes attention diffusion across irrelevant features~\cite{An2025AGLA}, obscuring instruction-critical fine-grained cues. (2) \textbf{Unfocused reasoning}. Both paradigms receive extensive historical contexts containing past observations and waypoints without prioritization. This hinders precise reasoning on high-value waypoints~\cite{zhou2024navgpt,chen2024mapgpt,10657191,chen2025affordances}. Long trajectory histories dilute attention, preventing models from isolating key cues for accurate reasoning~\cite{liu2024lostinthemiddle}.

These \emph{limitations} highlight two critical requirements: proactive acquisition of task-relevant visual observations to reduce perception redundancy, and focused reasoning over high-value waypoints within extensive historical contexts.

\begin{figure*}[t]
  \centering
  \includegraphics[width=1\linewidth]{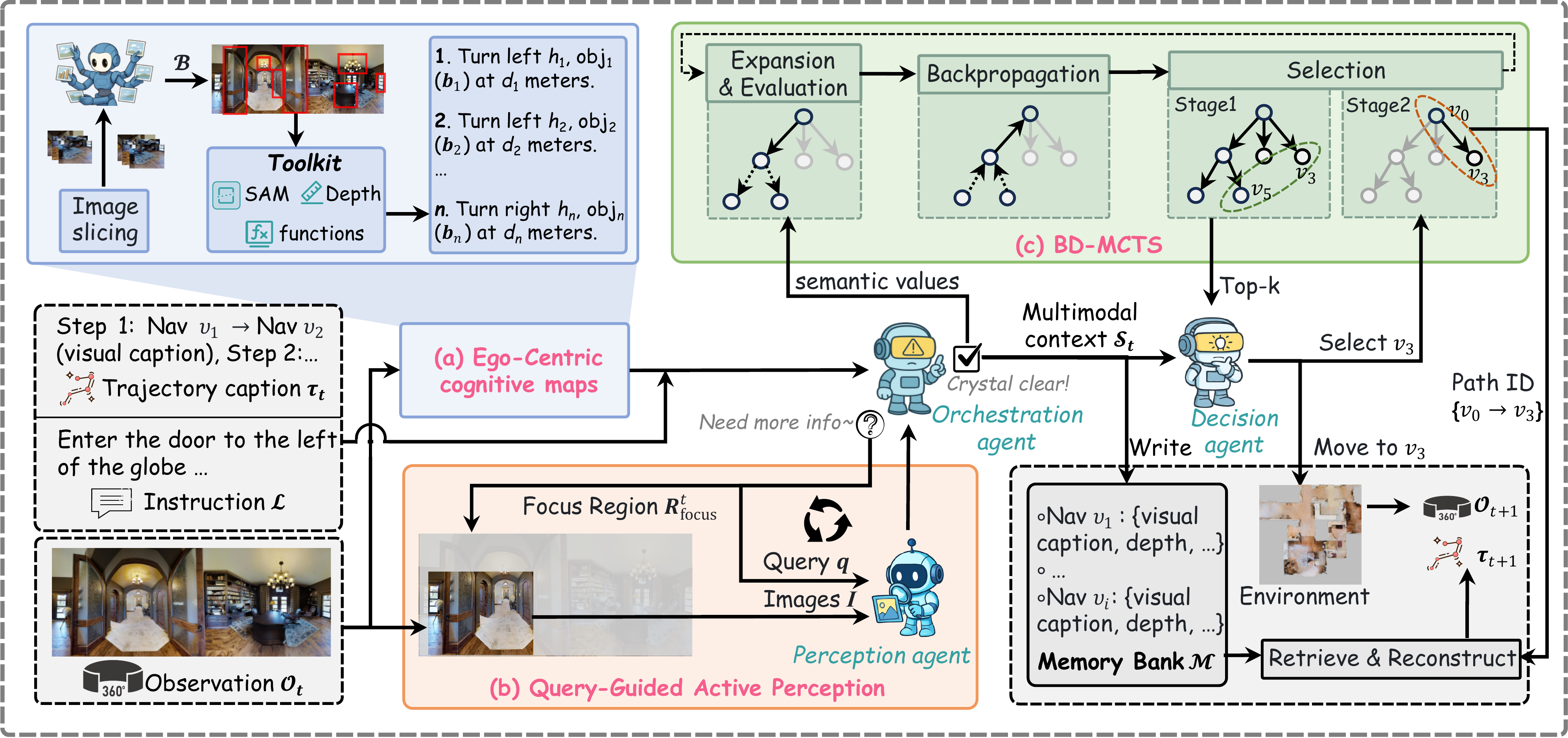}
  \caption{\textbf{Overview of the ProFocus framework.} 	ProFocus consists of two core mechanisms: 
  	\textit{Proactive Perception}, which selectively acquires instruction-relevant visual evidence through a closed perception--reasoning loop, 
  	and \textit{Focused Reasoning}, which leverages BD-MCTS to highlight top-$k$ high-value waypoints for effective focused reasoning.}
  \label{fig:architecture}
  \vspace{-3mm}
\end{figure*}

To address these challenges, we propose ProFocus, a training-free framework that unifies proactive perception and focused reasoning for efficient vision-and-language navigation. This is achieved through two core mechanisms, as illustrated in Figure~\ref{fig:architecture}. First, ProFocus introduces a reasoning-guided active perception mechanism to proactively acquire informative visual observations and assess the semantic values of navigable waypoints. It transforms panoramic observations into structured ego-centric semantic maps encoding object bounding boxes, depth, and directional relationships. Based on these maps, the orchestration agent (\ie, LLM) performs spatial reasoning to infer missing visual cues and generate targeted visual queries with corresponding focused regions. The perception agent (\ie, VLM) conducts fine-grained sensing exclusively within these regions. The orchestration agent iteratively evaluates information sufficiency and generates additional queries until adequate evidence is gathered, forming a closed perception-reasoning loop. Once sufficient information is acquired, the orchestration agent evaluates the semantic values of newly navigable waypoints to support subsequent focused reasoning. Furthermore, we design Branch-Diverse Monte Carlo Tree Search (BD-MCTS) to address spatial navigation uncertainty in unknown environments. Unlike standard MCTS that selects a single optimal action~\cite{browne2012survey,xu2025towards}, BD-MCTS identifies top-$k$ high-value candidates from extensive historical waypoints through path-aggregated value estimation, dynamic backpropagation, and branch-diversity constraints. The decision agent (\ie, LLM) then performs focused reasoning over these filtered candidates, achieving efficient global navigation.

The main contributions are summarized as follows:
\begin{itemize}
	\item A training-free progressive framework, ProFocus, that unifies proactive perception and focused reasoning for efficient vision-and-language navigation.
	\item A reasoning-guided proactive perception mechanism that establishes a closed perception-reasoning loop for iterative acquisition of instruction-relevant observations, rather than passively processing entire panoramic inputs.
	\item Branch-Diverse MCTS (BD-MCTS) that identifies top-$k$ high-value waypoints from extensive historical candidates, guiding the decision agent to focus on the contexts associated with these waypoints rather than treating all historical contexts indiscriminately.
\end{itemize}

\section{Related Work}
\label{sec:related}

\subsection{Vision-and-Language Navigation}
Benefiting from deep learning technologies~\cite{yang2025improvingmsa,yang2024pediatricsgpt,yang2025improving,yang2025medaide,liu2025reinforcement,lin2025sail,xue2023progressive,xue2024sagn,yang2022disentangled,yang2022emotion,yang2023context,yang2024robust,yang2024asynchronous,yang2024towards,yang2024MCIS}, early methods enhanced VLN via learning~\cite{anderson2018vision,zhu2020vision,hao2020towards,chen2021history,chen2022think,hong2021vln} and data augmentation~\cite{fried2018speaker,tan2019learning}. Traditional training-based methods, such as Speaker-Follower~\cite{fried2018speaker}, EnvDrop~\cite{tan2019learning}, and PREVALENT~\cite{hao2020towards}, employed Seq2Seq~\cite{sutskever2014sequence} or Transformer-based architectures~\cite{Vaswani2017}. However, scarcity of instructions and limited scene diversity hindered robust navigation policies, resulting in poor generalization.

Foundation models have introduced two paradigms for VLN~\cite{qin2025survey,chen2025affordances,zhang2024visionandlanguage,song2025towards}. The first centers on post-training adaptation~\cite{10657191,lin2025navcot,yu2025correctnav,liu2025nav,zhang2025cross}, where models are fine-tuned on domain-specific datasets. For instance, NaviLLM~\cite{10657191} unifies diverse navigation tasks through schema-based instructions and multi-view fusion. These methods achieve strong in-domain results but require significant computation and struggle with generalization. The second paradigm explores training-free methods~\cite{zhou2024navgpt,chen2025affordances,chen2024mapgpt,qiao2025open,long2024discuss} exploiting zero-shot reasoning without task-specific tuning. NavGPT~\cite{zhou2024navgpt} converts panoramic scenes into textual descriptions for GPT-4-based action prediction, while AO-Planner~\cite{chen2025affordances} employs visual affordance prompts using SAM for waypoint planning.

\subsection{Perception and Contextual Reasoning}
Despite the remarkable progress brought by foundation model-based VLN paradigms~\cite{zhang2025embodied,song2025towards}, these existing approaches~\cite{zhou2024navgpt,yu2025correctnav} still encounter two major challenges in perception and contextual reasoning. 
\textbf{(1) Perception.} Most existing methods~\cite{yu2025correctnav,zhang2025embodied,song2025towards} rely heavily on extensive visual inputs, such as panoramic observations~\cite{zhou2024navgpt,yu2025correctnav}, multi-view images~\cite{10657191,zhang2025embodied,song2025towards}, or continuous video streams~\cite{zhang2024uni}. For example, NaviLLM~\cite{10657191} processes multi-view inputs via Vision Transformers and fuses them for holistic scene understanding. While such designs capture rich environmental information, they uniformly process dense visual tokens, resulting in visual token inflation. This passive perception paradigm suggests the need for proactive acquisition that selectively gathers instruction-relevant observations.
\textbf{(2) Contextual Reasoning.} Existing methods often exhibit unfocused reasoning by treating historical contexts indiscriminately~\cite{zhou2024navgpt,chen2025affordances,yu2025correctnav}. They typically store all past information in the form of historical frames~\cite{yu2025correctnav,10657191}, textual descriptions~\cite{zhou2024navgpt,chen2025affordances}, or graph representations~\cite{chen2024mapgpt,zhang2025cross}. Although approaches such as NavFoM's~\cite{zhang2025embodied} token-based sampling and OpenNav's~\cite{qiao2025open} multi-granularity dynamic memory (MGDM) attempt to manage memory, they still rely on recency-based heuristics or fixed policies, failing to dynamically assess the semantic importance of waypoints. For instance, NavFoM's~\cite{zhang2025embodied} BATS allocates tokens based on temporal proximity rather than semantic relevance, hindering retrieval of critical waypoints during backtracking. These limitations motivate ProFocus, which introduces value-driven waypoint prioritization to enable focused reasoning over historical contexts.

\section{Methodology}

\subsection{Overview}
\label{sec:overview}

\noindent\textbf{Problem formulation.}
We formalize VLN over a navigation graph $\mathcal{G} = (\mathcal{V}, \mathcal{E})$ with waypoint set $\mathcal{V}$ and edge set $\mathcal{E} \subseteq \mathcal{V} \times \mathcal{V}$. Given a natural language instruction $\mathcal{I} \in \mathcal{L}$, the agent navigates from an initial waypoint through $\mathcal{G}$ discovered online. At timestep $t$, the agent at waypoint $v_t$ observes panoramic RGB image $\boldsymbol{\mathcal{O}}_t \in \mathbb{R}^{H \times W \times 3}$ and selects next waypoint $v_{t+1}$ from navigable candidates $\mathcal{N}_t = \{v \in \mathcal{V} : (v_t, v) \in \mathcal{E}\}$. A memory bank $\mathcal{M} = \{\mathcal{S}_0, \mathcal{S}_1, \ldots, \mathcal{S}_t\}$ stores multimodal contexts where each $\mathcal{S}_t$ records instruction, semantic map, and textual observations. A trajectory caption $\boldsymbol{\tau}_t $ constructed from $\mathcal{M}$ summarizes navigation history. Successful navigation requires perceiving $\boldsymbol{\mathcal{O}}_t$, reasoning over $\mathcal{I}$, $\mathcal{M}$ and $\boldsymbol{\tau}_t$, and selecting $v_{t+1}$ from $\mathcal{N}_t$.

\noindent\textbf{ProFocus framework.}
Rather than passively consuming predetermined visual inputs or uniformly treating all historical contexts, ProFocus addresses these challenges through a training-free framework unifying proactive perception and focused reasoning. The framework employs three specialized agents: orchestration agent $\mathcal{A}_{\mathrm{orch}}^{\boldsymbol{\theta}}$ (LLM for spatial inference and semantic evaluation), perception agent $\mathcal{A}_{\mathrm{perc}}^{\boldsymbol{\phi}}$ (VLM for fine-grained sensing), and decision agent $\mathcal{A}_{\mathrm{dec}}^{\boldsymbol{\psi}}$ (LLM for reasoning over top-$k$ candidates). As illustrated in Figure~\ref{fig:architecture}, ProFocus operates through two core mechanisms: Reasoning-Guided Proactive Perception (\S\ref{sec:proactive}) transforms panoramic observations into structured semantic maps (Figure~\ref{fig:architecture}(a)) and establishes a closed perception-reasoning loop to iteratively acquire instruction-relevant observations (Figure~\ref{fig:architecture}(b)). Focused Reasoning via Branch-Diverse MCTS (\S\ref{sec:focused}) maintains a global search tree to identify top-$k$ high-value waypoints from extensive historical contexts, enabling concentrated reasoning over strategically filtered candidates (Figure~\ref{fig:architecture}(c)).

\subsection{Reasoning-Guided Proactive Perception}
\label{sec:proactive}
Current methods process token-heavy panoramic views obscuring fine-grained cues~\cite{yu2025correctnav,zhang2025embodied} or rely on pre-captured descriptions omitting attributes~\cite{zhou2024navgpt}. Both exhibit passive perception, accepting predetermined inputs rather than actively querying instruction-relevant information. We establish a closed perception-reasoning loop enabling agents to determine \emph{``what to perceive''} based on \emph{``what is needed for decision-making.''} This offers three advantages: (\emph{i}) \textbf{reduced visual tokens} by focusing on instruction-relevant regions. (\emph{ii}) \textbf{enhanced attribute recognition} through targeted queries capturing fine-grained details (\eg, object color, texture, and spatial relations). (\emph{iii}) \textbf{adaptive perception} adjusting to navigation requirements.

\noindent\textbf{Ego-centric Semantic Map.}
As illustrated in Figure~\ref{fig:architecture}(a), given a panoramic observation $\mathcal{O}_t \in \mathbb{R}^{H \times W \times 3}$ at timestep $t$, we divide it into $K$ directional views $\{\mathcal{I}_k\}_{k=1}^K$, each with overlapping fields of view. A vision-language model $\mathit{F}_{\text{vlm}}^{\text{scan}}: \mathbb{R}^{H \times W \times 3} \to \mathcal{P}(\mathbb{R}^4 \times \mathcal{V})$ is then used to process these views in parallel to detect all objects present in the scene:
\begin{equation}
	\{(\boldsymbol{b}_i, \textit{obj}_i)\}_{i=1}^{N_t} = \mathit{F}_{\text{vlm}}^{\text{scan}}(\{\mathcal{I}_k\}_{k=1}^K),
\end{equation}
where $\boldsymbol{b}_i = (x_1, y_1, x_2, y_2) \in \mathbb{R}^4$ represents the bounding box of $i$, and $\textit{obj}_i \in \mathcal{V}$ denotes its category. To obtain depth for objects, we use monocular depth estimation~\cite{Bochkovskii2024:arxiv}, $d_i \leftarrow f_{\text{depth}}(\mathcal{O}_t, \boldsymbol{m}_i)$, where $\boldsymbol{m}_i$ is the object mask obtained using SAM2~\cite{ravi2024sam}. The heading angle of objects is computed as:
\begin{equation}
	h_i = \pi \cdot\left(\frac{x_1 + x_2 - F}{F}\right) \in [-\pi, \pi],
\end{equation}
where $F$ is the field of view width. Finally, the ego-centric semantic map is constructed as:
{\small
\begin{equation}
		\mathcal{C}_t = \{(h_i, \textit{obj}_i(\boldsymbol{b}_i), d_i)\}_{i=1}^{N_t},
\end{equation}
}%
which is formatted as natural language text, such as \textit{``turn left/right $h_i$ degrees, $\textit{obj}_i$ (bounding box $\boldsymbol{b}_i$) at $d_i$ meters.''} This structured map allows LLMs to reason about object locations (\eg, "objects to the left"), enabling spatial understanding and focused region generation.


\noindent\textbf{Reasoning-Driven Perception Loop.}
As shown in Figure~\ref{fig:architecture}(b), given the semantic map $\mathcal{C}_t$, trajectory history $\boldsymbol{\tau}_t$, and instruction $\mathcal{I}$, the orchestration agent $\mathcal{A}_{\mathrm{orch}}^{\boldsymbol{\theta}}$ initiates an active perception loop by generating a visual query $\boldsymbol{q} \in \mathcal{L}$ paired with a focused region $\boldsymbol{R}_{\text{focus}}^t \subseteq [0,H] \times [0,W]$:
\begin{equation}
	\label{eq:query_gen}
	(\boldsymbol{q}, \boldsymbol{R}_{\text{focus}}^t) = \mathcal{A}_{\mathrm{orch}}^{\boldsymbol{\theta}}(\mathcal{C}_t, \boldsymbol{\tau}_t, \mathcal{I}, \mathcal{H}_{\text{query}}),
\end{equation}
where $\mathcal{H}_{\text{query}} = \{(\boldsymbol{q}_j, \boldsymbol{a}_j^t)\}_{j=1}^{i-1}$ maintains the history of query-response pairs from previous iterations, preventing redundant queries. The perception agent $\mathcal{A}_{\mathrm{perc}}^{\boldsymbol{\phi}}$ performs fine-grained analysis within the focused region $\boldsymbol{R}_{\text{focus}}^t$:
\begin{equation}
	\label{eq:perception}
	\boldsymbol{a}_i^t = \mathcal{A}_{\mathrm{perc}}^{\boldsymbol{\phi}}(\boldsymbol{\mathcal{O}}_t|_{\boldsymbol{R}_{\text{focus}}^t}, \boldsymbol{q}),
\end{equation}
where $\boldsymbol{\mathcal{O}}_t|_{\boldsymbol{R}_{\text{focus}}^t}$ denotes the cropped region of the panoramic observation $\boldsymbol{\mathcal{O}}_t$, and $\boldsymbol{a}_i^t \in \mathcal{L}$ represents the textual description of visual attributes extracted from the $i$-th query, including object color, texture, and spatial relationships. After each query and perception step, the orchestration agent evaluates whether the collected information is sufficient:
\begin{equation}
	\label{eq:info_sufficiency}
	s_t = \mathcal{A}_{\mathrm{orch}}^{\boldsymbol{\theta}}(\mathcal{C}_t, \mathcal{H}_{\text{query}}, \mathcal{I}), \quad s_t \in \{\text{sufficient}, \text{insufficient}\}.
\end{equation}
If $s_t = \text{insufficient}$, the orchestration agent generates queries and focused regions for the perception agent, continuing the closed loop. This iterative cycle accumulates descriptions $\{\boldsymbol{a}_i^t\}_{i=1}^{n_t}$ from $n_t$ iterations until the information is deemed sufficient. Once the perception loop terminates, the orchestration agent evaluates the semantic values of newly discovered waypoints $\mathcal{A}_{\text{new}}$ by leveraging all accumulated visual information:
{\small
\begin{equation}
\label{eq:sem_value}
		V_{\text{sem}}(u) = \mathcal{A}_{\mathrm{orch}}^{\boldsymbol{\theta}}(\mathcal{I}, \boldsymbol{\tau}_t, \{\boldsymbol{a}_i^t\}_{i=1}^{n_t}), \quad V_{\text{sem}}(u) \in [0, 1],
	\end{equation}
}%
where $\{\boldsymbol{a}_i^t\}_{i=1}^{n_t}$ aggregates descriptions from perception loop, and $V_{\text{sem}}(u)$ quantifies how well waypoint $u$ aligns with the navigation instruction $\mathcal{I}$ given trajectory history $\boldsymbol{\tau}_t$ and accumulated observations. Higher values indicate stronger semantic relevance to navigation goals, providing the foundation for value backpropagation in BD-MCTS (\S\ref{sec:focused}). Finally, the accumulated textual information is integrated with navigation instruction $\mathcal{I}$, trajectory history $\boldsymbol{\tau}_t$, and semantic map $\mathcal{C}_t$ to form the multimodal context $\mathcal{S}_t$:
{\small
	\begin{equation}
		\label{eq:multimodal_context}
		\mathcal{S}_t = \{\mathcal{I}, \boldsymbol{\tau}_t, \mathcal{C}_t, \{\boldsymbol{a}_i^t\}_{i=1}^{n_t}\},
\end{equation}
}%
where $\mathcal{C}_t$ encodes the semantic map with object locations. This unified textual representation $\mathcal{S}_t$ is stored in the memory bank $\mathcal{M}$ and enables the decision agent to perform global reasoning over complete navigation context.

\subsection{Focused Reasoning via Branch-Diverse MCTS}
\label{sec:focused}
As navigation progresses, the memory bank $\mathcal{M} = \{\mathcal{S}_0, \mathcal{S}_1, \ldots, \mathcal{S}_t\}$ accumulates extensive multimodal contexts. The decision agent $\mathcal{A}_{\mathrm{dec}}^{\boldsymbol{\psi}}$ must reason over all accumulated contexts without prioritization, diffusing attention uniformly and leading to inefficient decision-making. Humans do not uniformly evaluate all past states when planning. Instead they \emph{prune} low-value branches and use \emph{prioritized memory access} to selectively replay task-relevant trajectories given current goals~\cite{huys2012bonsai,mattar2018prioritized}. Inspired by this, we propose Branch-Diverse MCTS (BD-MCTS), as illustrated in Figure~\ref{fig:architecture}(c). BD-MCTS filters a compact top-$k$ set of high-value waypoints and iteratively refines their values with new observations. This enables the decision agent to focus reasoning on these candidates rather than diffusing attention across all historical contexts.

\noindent\textbf{Tree-Graph Adaptation.}
MCTS assumes acyclic state spaces, but navigation graphs contain cycles. We maintain search tree $\mathcal{T} = \langle \boldsymbol{V}_{\mathcal{T}}, \boldsymbol{E}_{\mathcal{T}}, Q, N \rangle$ where $\boldsymbol{V}_{\mathcal{T}}$ are discovered waypoints, $\boldsymbol{E}_{\mathcal{T}}$ are explored edges, $Q$ maps waypoints to estimated quality, and $N$ tracks visit counts. When expanding at $v_t$, new waypoints are added as children while cross-branch waypoints are skipped to preserve acyclicity.

\noindent\textbf{Phase I: Expansion with Semantic Evaluation.}
Classical MCTS expands new nodes and estimates their values through random rollouts. We replace rollouts with semantic value estimation. At current waypoint $v_t$, we discover new navigable candidates $\mathcal{A}_{\text{new}}$ and expand the tree:
{\small
\begin{equation}
		\label{eq:bd_expansion}
		\boldsymbol{V}_{\mathcal{T}} \leftarrow \boldsymbol{V}_{\mathcal{T}} \cup \mathcal{A}_{\text{new}}, \quad \boldsymbol{E}_{\mathcal{T}} \leftarrow \boldsymbol{E}_{\mathcal{T}} \cup \{(v_t, u) : u \in \mathcal{A}_{\text{new}}\}.
\end{equation}
}%
Each expanded waypoint $u \in \mathcal{A}_{\text{new}}$ is initialized with the semantic value $V_{\text{sem}}(u)$ computed via Eq.~\ref{eq:sem_value}, which integrates instruction $\mathcal{I}$, trajectory history $\boldsymbol{\tau}_t$, and accumulated visual observations $\{\boldsymbol{a}_i^t\}_{i=1}^{n_t}$ from proactive perception. This provides value estimates grounded in multimodal context:
\begin{equation}
	\label{eq:bd_init}
	Q(u) \leftarrow V_{\text{sem}}(u), \quad N(u) \leftarrow 0, \quad \forall u \in \mathcal{A}_{\text{new}}.
\end{equation}
This unified phase enables semantically grounded initialization while avoiding computational cost of rollouts.

\noindent\textbf{Phase II: Backpropagation with Dynamic Refinement.}
Classical MCTS backpropagates rollout results to update ancestor statistics. We backpropagate semantic values to dynamic refinement. We compute reward $R_t$ as the average semantic value of newly expanded waypoints when $|\mathcal{A}_{\text{new}}| > 0$, otherwise using current quality $Q(v_t)$. This reward propagates along root-to-current path $\boldsymbol{\pi}(r, v_t)$:
{\small
\begin{equation}
		\label{eq:bd_update}
		N(v) \leftarrow N(v) + 1; \quad Q(v) \leftarrow Q(v) + \frac{R_t - Q(v)}{N(v)}.
\end{equation}
}%
This asymmetric backpropagation updates only the selected path, creating priority shifts. Low reward triggers backtracking while high reward reinforces forward exploration.

\noindent\textbf{Phase III: Top-$k$ Selection with Branch Diversity.}
After iterative expansion and backpropagation cycles, the search tree $\mathcal{T}$ accumulates value estimates across all explored waypoints. We distill the extensive tree into a compact set of top-$k$ high-value candidates to guide the decision agent's focused reasoning. For each leaf node $v \in \mathcal{L}$, we evaluate its comprehensive quality by aggregating backpropagated values along its root-to-leaf path $\boldsymbol{\pi}(v) = [v_0=r, v_1, \ldots, v_L=v]$. A leaf's trustworthiness depends not only on its own semantic value but also on the reliability of the trajectory leading to it. We compute path-aggregated value by weighting each ancestor's quality $Q(v_i)$ by its normalized visit count, giving higher weight to frequently visited nodes representing verified navigation segments while allowing newly discovered leaves to contribute through their initial semantic values. This balances exploitation of reliable paths with exploration of novel discoveries. However, ranking by aggregated value may yield spatially distant waypoints that are impractical to navigate. To incorporate physical reachability constraints, we apply a distance penalty based on shortest path length $d_{\mathcal{G}}(v_t, v)$ in graph $\mathcal{G}$, normalized by maximum distance. The final scoring function combines path value with distance penalty:
\begin{equation}
	\label{eq:bd_score}
	\text{Score}(v) = V_{\text{path}}(v) - \lambda \cdot \frac{d_{\mathcal{G}}(v_t, v)}{\max_{u \in \mathcal{L}} d_{\mathcal{G}}(v_t, u)},
\end{equation}
where $V_{\text{path}}(v)$ denotes the visit-count weighted path-aggregated value, and $\lambda$ controls penalty strength. Finally, to ensure strategic diversity spanning different exploration directions, we select top-$k$ leaves from $\mathcal{L}$ ranked by $\text{Score}(v)$, with the constraint that each parent node contributes at most 2 children to the candidate set $\mathcal{C}_k$. 

The decision agent $\mathcal{A}_{\mathrm{dec}}^{\boldsymbol{\psi}}$ performs fine-grained reasoning over the filtered candidate set $\mathcal{P}_{new}$:
\begin{equation}
	\label{eq:bd_decision}
	v_{t+1} = \mathcal{A}_{\mathrm{dec}}^{\boldsymbol{\psi}}\left(\mathcal{P}_{new}, \mathcal{S}_t, \mathcal{M}\right),
\end{equation}
where detailed historical context for each candidate is retrieved from memory bank $\mathcal{M}$ by indexing waypoint identifiers along path $\boldsymbol{\pi}(r, v)$, and current multimodal context $\mathcal{S}_t$ provides instruction, trajectory, and observations. This two-stage mechanism balances systematic global exploration with focused decision-making. BD-MCTS performs coarse-grained filtering from extensive history to top-$k$ candidates, followed by LLM performing nuanced contextual reasoning. This avoids both premature commitment and attention diffusion. After selecting $v_{t+1}$, the agent transitions to the new waypoint, updates memory $\mathcal{M} \leftarrow \mathcal{M} \cup \{\mathcal{S}_{t+1}\}$. 
It then reconstructs the trajectory caption
$\boldsymbol{\tau}_{t+1} \leftarrow f_{\text{rec}}(\{v_0, \ldots, v_{t+1}\}, \mathcal{M})$, where $f_{\text{rec}}$ is the retrieve-and-reconstruct function. The agent obtains the  observation $\boldsymbol{\mathcal{O}}_{t+1}$ and repeats this cycle until reaching the target.

\section{Experiments}

\subsection{Experimental Setup}

\textbf{Datasets.} Experiments are conducted on two widely-used VLN benchmarks: R2R~\cite{anderson2018vision} and REVERIE~\cite{qi2020reverie}. R2R requires agents to follow detailed step-by-step instructions to reach target locations, while REVERIE combines navigation with object grounding, requiring agents to locate specific objects based on high-level descriptions. Following recent foundation model-based VLN works~\cite{zhou2024navgpt,chen2024mapgpt}, we focus solely on navigation for REVERIE. We evaluate whether the agent successfully reaches the target location, without assessing object grounding.

\noindent\textbf{Evaluation Metrics.} Standard evaluation metrics~\cite{anderson2018vision} are adopted: (1) \textbf{Success Rate (SR)}: percentage of episodes stopping within 3m of target. (2) \textbf{Oracle Success Rate (OSR)}: SR with oracle stop policy. (3) \textbf{Navigation Error (NE)}: average distance to target in meters. (4) \textbf{Success weighted by Path Length (SPL)}: SR penalized by path efficiency. Following the REVERIE benchmark convention~\cite{chen2024mapgpt}, we report OSR, SR, and SPL but omit NE, as the task prioritizes object-centric navigation success over precise distance measurement.

\noindent\textbf{Implementation Details.} 
Our training-free framework consists of three specialized agents: an orchestration agent, a perception agent, and a decision agent, powered by LLM, VLM, and LLM respectively. We select state-of-the-art large language models Qwen3-Max~\cite{yang2025qwen3} and DeepSeek-V3~\cite{deepseekai2024deepseekv3}, and vision-language models Qwen3-VL-Max~\cite{yang2025qwen3} and GLM-4.5V~\cite{hong2025glm} for evaluation. We denote these models as Q3, DS3, Q3VL, and GLM respectively for brevity. We evaluate two model configurations for ProFocus: (1) DeepSeek-V3 + GLM-4.5V. (2) Qwen3-Max + Qwen3-VL-Max. To enable fair comparison across different foundation models, following~\cite{chen2024mapgpt}, we re-implement NavGPT with LLM backbones Q3 and DS3, and MapGPT with VLM backbones Q3VL and GLM on a randomly sampled validation subset under identical experimental settings. These re-implemented results are marked as $^\dagger$ in Tables~\ref{tab:r2r_results} and~\ref{tab:reverie_results}.

\begin{table}[t]
\centering
\small
\setlength{\tabcolsep}{4.5pt}  
\renewcommand{\arraystretch}{1.15}
\caption{Results on R2R validation unseen set. $^\dagger$ denotes our re-implementation with different foundation models. \textbf{Bold} best, \underline{underline} second-best \emph{within each method group}.}
\label{tab:r2r_results}
\begin{tabularx}{\linewidth}{Xcccc}
\toprule
\textbf{Methods} & \textbf{NE}\,$\downarrow$ & \textbf{OSR}\,$\uparrow$ & \textbf{SR}\,$\uparrow$ & \textbf{SPL}\,$\uparrow$ \\
\midrule

\rowcolor{GroupBlue}
\multicolumn{5}{l}{\textbf{Training-based VLN specialists}} \\[-2pt]
Seq2Seq~\cite{anderson2018vision}            & 7.81 & \underline{28}   & \underline{21}   & -- \\
Speaker-Follower~\cite{fried2018speaker}      & \underline{6.62} & \textbf{45}   & 36   & -- \\
EnvDrop~\cite{tan2019learning}                & \textbf{5.22} & --   & \textbf{52}   & \textbf{48} \\
\midrule

\rowcolor{GroupBlue}
\multicolumn{5}{l}{\textbf{Pretraining-based VLN specialists}} \\[-2pt]
PREVALENT~\cite{hao2020towards}               & 4.71 & --   & 58   & 53 \\
HAMT~\cite{chen2021history}                   & \textbf{2.29} & --   & 66   & \textbf{61} \\
DUET~\cite{chen2022think}                     & \underline{3.31} & \textbf{81}   & \textbf{72}   & \underline{60} \\
NavCoT~\cite{lin2025navcot}                   & 6.26 & \underline{48}   & 40   & 37 \\
NaviLLM~\cite{10657191}               & 3.51 & --   & \underline{67}   & 59 \\
\midrule

\rowcolor{GroupBlue}
\multicolumn{5}{l}{\textbf{Zero-shot VLN with Foundation Models}} \\[-2pt]
NavGPT (GPT-4)~\cite{zhou2024navgpt}          & 6.46 & 42.0 & 34.0 & 29.0 \\
MapGPT (GPT-4)~\cite{chen2024mapgpt}          & 6.29 & 57.6 & 38.8 & 25.8 \\
MapGPT (GPT-4V)~\cite{chen2024mapgpt}         & 5.63 & 57.6 & 43.7 & 34.8 \\
DiscussNav (GPT-4)~\cite{long2024discuss}     & 5.32 & 61.0 & 43.0 & \underline{40.0} \\
MSNav (GPT-4o)~\cite{liu2025msnav}            & 5.24 & \underline{65.0} & 46.0 & \underline{40.0} \\
NavGPT$^\dagger$ (DS3)                        & 6.52 & 50.0 & 36.0 & 28.1 \\
NavGPT$^\dagger$ (Q3)                         & \textbf{4.82} & 57.5 & 47.0 & 38.4 \\
MapGPT$^\dagger$ (GLM)                        & 5.00 & \textbf{70.7} & 41.4 & 30.8 \\
MapGPT$^\dagger$ (Q3VL)                       & 5.27 & 47.5 & 32.0 & 28.7 \\
\textbf{Ours$^\dagger$ (DS3+GLM)}             & 5.21 & 63.0 & \underline{50.0} & \textbf{41.2} \\
\textbf{Ours$^\dagger$ (Q3+Q3VL)}             & \underline{4.92} & \underline{65.0} & \textbf{52.5} & 39.8 \\
\bottomrule
\end{tabularx}
\vspace{-3mm}
\end{table}

\begin{table}[t]
\centering
\small
\setlength{\tabcolsep}{8pt}
\renewcommand{\arraystretch}{1.05}
\caption{Results on REVERIE validation unseen set. $^\dagger$ denotes our re-implementation with different foundation models. \textbf{Bold} best, \underline{underline} second-best \emph{within each method group}.}
\label{tab:reverie_results}
\begin{tabularx}{\linewidth}{Xccc}
\toprule
\textbf{Methods} & \textbf{OSR}\,$\uparrow$ & \textbf{SR}\,$\uparrow$ & \textbf{SPL}\,$\uparrow$ \\
\midrule

\rowcolor{GroupBlue}
\multicolumn{4}{l}{\textbf{Training-based VLN specialists}} \\[-2pt]
Seq2Seq~\cite{anderson2018vision}            & 8.1  & 4.2  & 2.8 \\
FAST-MATTN~\cite{qi2020reverie}              & \underline{28.2} & \underline{14.4} & \underline{7.2} \\
NavQ~\cite{xu2025navq}                       & \textbf{62.0} & \textbf{54.1} & \textbf{39.2} \\
\midrule

\rowcolor{GroupBlue}
\multicolumn{4}{l}{\textbf{Pretraining-based VLN specialists}} \\[-2pt]
HAMT~\cite{chen2021history}                  & 35.4 & 31.6 & 29.6 \\
DUET~\cite{chen2022think}                    & \underline{50.0} & \textbf{45.8} & \underline{35.3} \\
NaviLLM~\cite{10657191}              & \textbf{53.7} & \underline{44.6} & \textbf{36.6} \\
\midrule

\rowcolor{GroupBlue}
\multicolumn{4}{l}{\textbf{Zero-shot VLN with Foundation Models}} \\[-2pt]
NavGPT (GPT-4)~\cite{zhou2024navgpt}         & 28.3 & 19.2 & 14.6 \\
MapGPT (GPT-4)~\cite{chen2024mapgpt}         & 42.6 & 28.4 & 14.5 \\
MapGPT (GPT-4V)~\cite{chen2024mapgpt}        & 36.8 & 31.6 & 20.3 \\
NavGPT$^\dagger$ (DS3)                       & 34.0 & 21.0 & 14.3 \\
NavGPT$^\dagger$ (Q3)                        & 24.0 & 18.0 & 13.7 \\
MapGPT$^\dagger$ (GLM)                       & 50.5 & 37.1 & 24.7 \\
MapGPT$^\dagger$ (Q3VL)                      & 36.5 & 22.5 & 18.0 \\
\textbf{Ours$^\dagger$ (DS3+GLM)}            & \textbf{57.1} & \underline{36.9} & \textbf{25.9} \\
\textbf{Ours$^\dagger$ (Q3+Q3VL)}            & \underline{51.7} & \textbf{40.0} & \underline{24.8} \\
\bottomrule
\end{tabularx}
\vspace{-3mm}
\end{table}

\subsection{Performance Analysis and Comparisons}
We analyze ProFocus's performance across four key dimensions: navigation capability, exploration-success trade-offs, long-trajectory robustness, and cross-dataset generalization.

\noindent\textbf{Superior Navigation Capability.}
ProFocus achieves state-of-the-art zero-shot performance on both SR and SPL metrics (Tables~\ref{tab:r2r_results} and~\ref{tab:reverie_results}). On R2R, ProFocus (Q3+Q3VL) reaches 52.5\% SR and 39.8\% SPL, substantially exceeding re-implemented baselines: NavGPT (Q3) at 47.0\% SR and 38.4\% SPL, MapGPT (Q3VL) at 32.0\% SR and 28.7\% SPL. The DS3+GLM configuration achieves 50.0\% SR and 41.2\% SPL, outperforming NavGPT (DS3) at 36.0\% SR and 28.1\% SPL, MapGPT (GLM) at 41.4\% SR and 30.8\% SPL. Navigation error remains competitive at 4.92m, comparable to NavGPT (Q3) at 4.82m. On REVERIE, ProFocus (Q3+Q3VL) achieves 40.0\% SR and 24.8\% SPL versus NavGPT (Q3) at 18.0\% SR and 13.7\% SPL, MapGPT (Q3VL) at 22.5\% SR and 18.0\% SPL. ProFocus (DS3+GLM) reaches 36.9\% SR and 25.9\% SPL versus MapGPT (GLM) at 37.1\% SR and 24.7\% SPL. The consistent improvements across SR and SPL demonstrate ProFocus's effectiveness in accurate navigation.

\noindent\textbf{Balanced Exploration and Success.}
The OSR-SR gap exhibits task-dependent characteristics. In most standard navigation scenarios, ProFocus demonstrates highly effective stopping mechanisms. On R2R, MapGPT (GLM) creates a 29.3\% gap between OSR and SR. ProFocus (Q3+Q3VL) significantly narrows this difference to 12.5\%. This shows our agent efficiently identifies goal arrival to prevent over-exploration. However, object-grounding tasks like REVERIE introduce inherent localization uncertainty. This specific requirement alters the OSR-SR dynamics. The gap may fluctuate relative to certain baselines in these specific cases. Despite these minor fluctuations, ProFocus maintains superior overall accuracy. On REVERIE, ProFocus (Q3+Q3VL) achieves a 40.0\% SR. ProFocus (DS3+GLM) reaches a 36.9\% SR. These results prove that ProFocus robustly converts exploration into high navigation success across diverse tasks.

\begin{figure}[t]
  \centering
  \includegraphics[width=\linewidth]{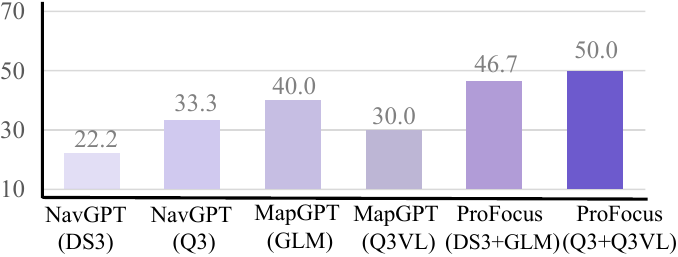}
  \caption{Performance comparison on the 30 longest navigation episodes from R2R validation unseen set.}
  \label{fig:long_trajectory}
  \vspace{-1mm}
\end{figure}

\noindent\textbf{Long-Trajectory Robustness.}
To evaluate robustness in challenging scenarios, we analyze the 30 longest navigation episodes from R2R validation unseen set (Figure~\ref{fig:long_trajectory}). These long trajectories amplify decision complexity, requiring agents to maintain coherent reasoning over dozens of historical waypoints. ProFocus demonstrates advantages in these conditions. ProFocus (Q3+Q3VL) achieves 50.0\% SR, exceeding MapGPT (GLM) at 40.0\%. NavGPT variants show weaker performance at 33.3\% and 22.2\% SR. ProFocus (DS3+GLM) reaches 46.7\% SR, consistently outperforming baselines. The performance advantage in long trajectories validates ProFocus's capability to effectively handle extensive historical contexts without performance degradation as trajectory complexity increases.

\noindent\textbf{Cross-Dataset and Cross-Model Consistency.}
ProFocus maintains robust performance across diverse settings. Both configurations consistently outperform corresponding baselines on R2R (detailed step-by-step instructions) and REVERIE (object-centric navigation). ProFocus (Q3+Q3VL) achieves the highest SR on both datasets: 52.5\% on R2R and 40.0\% on REVERIE among re-implemented methods. ProFocus (DS3+GLM) reaches 50.0\% SR on R2R and 36.9\% SR on REVERIE, maintaining competitive performance. ProFocus demonstrates consistent advantages across different foundation model combinations, substantially exceeding corresponding baselines on both datasets. This stability demonstrates that ProFocus's performance gains are consistent across different foundation model capabilities and dataset characteristics.

\begin{table}[t]
	\centering
	\small
	\setlength{\tabcolsep}{6pt}
	\renewcommand{\arraystretch}{1.15}
	\caption{Ablation study on R2R validation unseen set. BD-MCTS: Branch-Diverse MCTS for focused reasoning, PP: Proactive Perception mechanism.}
    \vspace{-2mm}
	\label{tab:model_comparison}
	\begin{tabular}{lcccc}
		\toprule
		\textbf{Methods} & \textbf{NE}\,$\downarrow$ & \textbf{OSR}\,$\uparrow$ & \textbf{SR}\,$\uparrow$ & \textbf{SPL}\,$\uparrow$ \\
		\midrule
		
	\rowcolor{GroupBlue}
	\multicolumn{5}{l}{\textbf{DS3+GLM}} \\
		ProFocus                             & 5.21 & 63.0 & 50.0 & 41.2 \\
		ProFocus w/o BD-MCTS                 & 5.75 & 54.2 & 46.9 & 40.0 \\
		ProFocus w/o PP                      & 6.45 & 50.0 & 42.0 & 30.0  \\
		\midrule
		
	\rowcolor{GroupBlue}
	\multicolumn{5}{l}{\textbf{Q3+Q3VL}} \\
	ProFocus                             & 4.92 & 65.0 & 52.0 & 39.8 \\
	ProFocus w/o BD-MCTS                 & 6.19 & 53.66 & 50.0 & 38.4 \\
	ProFocus w/o PP                      & 5.97 & 54.0 & 48.0 & 34.4 \\
	\bottomrule
\end{tabular}
\end{table}

\begin{table}[t]
	\centering
	\small
	\setlength{\tabcolsep}{8pt}
	\renewcommand{\arraystretch}{1.15}
	\caption{Ablation study on REVERIE validation unseen set. BD-MCTS: Branch-Diverse MCTS for focused reasoning, PP: Proactive Perception mechanism. }
    \vspace{-2mm}
	\label{tab:ablation_reverie}
	\begin{tabularx}{\linewidth}{Xccc}
		\toprule
		\textbf{Methods} & \textbf{OSR}\,$\uparrow$ & \textbf{SR}\,$\uparrow$ & \textbf{SPL}\,$\uparrow$ \\
		\midrule
		
	\rowcolor{GroupBlue}
	\multicolumn{4}{l}{\textbf{DS3+GLM}} \\
		ProFocus                              & 57.14 & 36.90 & 25.85 \\
		ProFocus w/o BD-MCTS                  & 42.67 & 33.33 & 26.73 \\
		ProFocus w/o PP                       & 36.00 & 32.50 & 23.05 \\
		\midrule
		
	\rowcolor{GroupBlue}
	\multicolumn{4}{l}{\textbf{Q3+Q3VL}} \\
	ProFocus                              & 51.67 & 40.00 & 24.83 \\
	ProFocus w/o BD-MCTS                  & 48.89 & 37.78 & 27.37 \\
	ProFocus w/o PP                       & 34.00 & 30.00 & 18.51 \\
	\bottomrule
\end{tabularx}
\vspace{-3mm}
\end{table}

\begin{figure*}[t]
  \centering
  \includegraphics[width=0.98\linewidth]{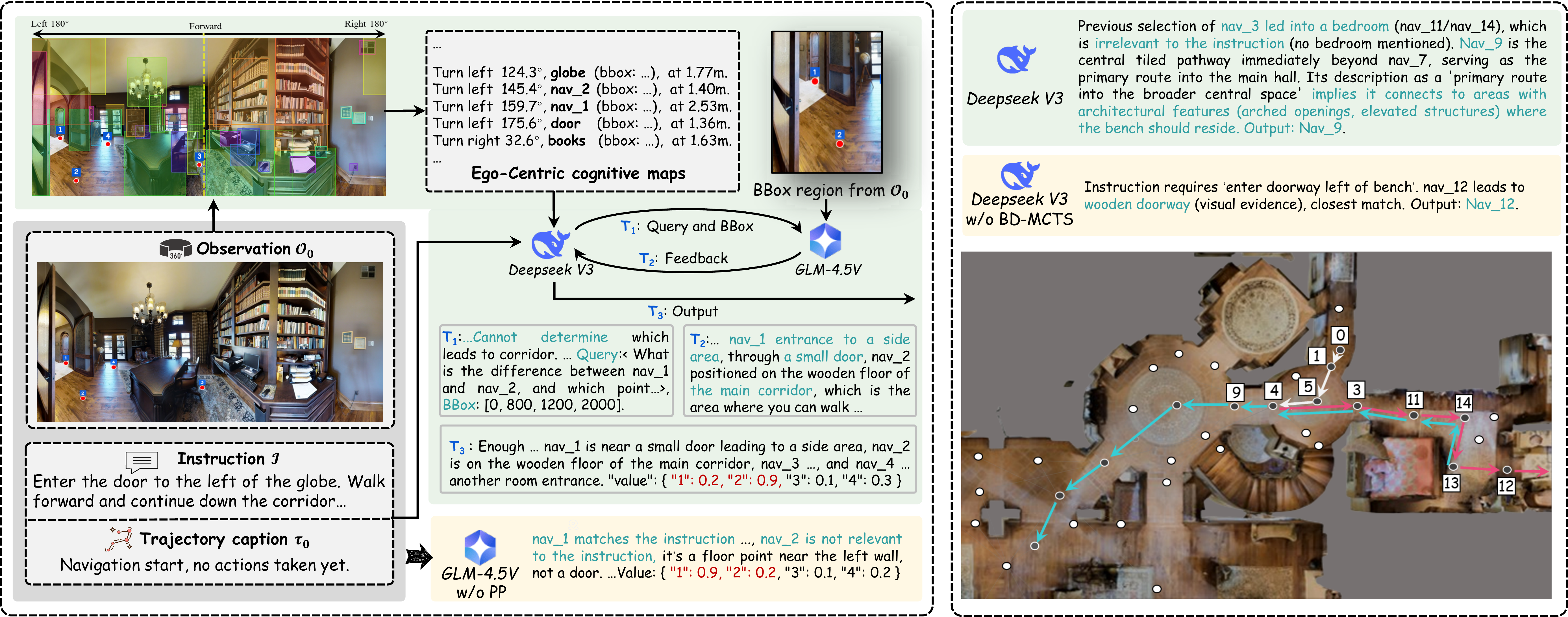}
  
  \caption{\textbf{Qualitative case study demonstrating proactive perception and focused reasoning.} \textit{Left}: Proactive perception enables accurate semantic value estimation by acquiring fine-grained visual details. \textit{Right}: BD-MCTS-guided top-$k$ focused reasoning enables error correction through global historical awareness.}
  \label{fig:qualitative_case}
  \vspace{-2mm}
\end{figure*}

\subsection{Ablation Studies}

We evaluate two ablation variants under two foundation model configurations (Tables~\ref{tab:model_comparison} and~\ref{tab:ablation_reverie}): (\emph{i}) ProFocus w/o BD-MCTS removes the focused reasoning mechanism, requiring the decision agent to reason over all historical waypoints without top-$k$ filtering. (\emph{ii}) ProFocus w/o PP removes the proactive perception mechanism, processing full panoramic views without targeted visual queries.

\noindent\textbf{Impact of BD-MCTS on Exploration Capability.}
Removing BD-MCTS causes OSR degradation across both datasets. On R2R (Table~\ref{tab:model_comparison}), OSR drops from 63.0\% to 54.2\% for DS3+GLM and from 65.0\% to 53.66\% for Q3+Q3VL. On REVERIE (Table~\ref{tab:ablation_reverie}), the impact is even more pronounced: OSR decreases from 57.14\% to 42.67\% for DS3+GLM and from 51.67\% to 48.89\% for Q3+Q3VL. In contrast, SR degradation is modest. The disproportionate OSR versus SR degradation suggests that without BD-MCTS, agents explore less effectively. They fail to discover target-proximate waypoints even when navigation succeeds. Navigation error increases substantially on R2R (4.92m to 6.19m for Q3+Q3VL), validating the importance of systematic historical context filtering for spatial reasoning.

\noindent\textbf{Impact of Proactive Perception on Path Efficiency.}
Removing proactive perception causes severe performance degradation across all metrics. SPL losses are particularly dramatic. On R2R (Table~\ref{tab:model_comparison}), SPL drops from 41.2\% to 30.0\% for DS3+GLM and from 39.8\% to 34.4\% for Q3+Q3VL. SR decreases from 50.0\% to 42.0\% for DS3+GLM and from 52.0\% to 48.0\% for Q3+Q3VL. On REVERIE (Table~\ref{tab:ablation_reverie}), OSR drops dramatically from 51.67\% to 34.0\% for Q3+Q3VL and from 57.14\% to 36.0\% for DS3+GLM. The substantial OSR and SPL degradation indicates that without targeted visual queries, agents struggle to acquire instruction-relevant spatial details. This leads to inefficient exploration and poor navigation decisions.

\noindent\textbf{Task-Specific Sensitivity.}
The two mechanisms exhibit different sensitivities across R2R and REVERIE. BD-MCTS removal causes greater OSR loss on REVERIE than R2R. For DS3+GLM, REVERIE shows 14.47\% decline versus 8.8\% on R2R. This suggests REVERIE's object-centric navigation demands more effective historical context management. Conversely, proactive perception removal causes greater SPL loss on R2R than REVERIE. For DS3+GLM, R2R shows 11.2\% absolute decline in SPL. This indicates R2R's detailed step-by-step instructions require more precise visual attribute acquisition for path efficiency. An interesting phenomenon emerges on REVERIE: removing BD-MCTS slightly increases SPL from 25.85\% to 26.73\% for DS3+GLM and from 24.83\% to 27.37\% for Q3+Q3VL, while reducing OSR and SR. This suggests that without top-$k$ filtering, agents adopt more conservative, shorter paths at the cost of exploration breadth and success rate.

\noindent\textbf{Cross-Model Consistency.}
Both mechanisms demonstrate consistent contributions across foundation model combinations. Removing BD-MCTS consistently reduces OSR across configurations and datasets. Removing proactive perception causes substantial losses in both OSR and SPL. On REVERIE, Q3+Q3VL shows greater sensitivity to proactive perception removal with OSR dropping 17.67\% compared to 21.14\% for DS3+GLM. Both configurations exhibit comparable degradation on R2R. This cross-model consistency validates that both mechanisms provide fundamental architectural improvements rather than compensating for specific foundation model weaknesses.

\subsection{Qualitative Analysis}
Figure~\ref{fig:qualitative_case} (left) illustrates the advantage of proactive perception in semantic value estimation. Without proactive perception (bottom, yellow), the agent processes the full panoramic view and identifies that waypoint 1 leads to a corridor. It assigns the highest value of 0.9 to waypoint 1 and only 0.2 to waypoint 2. However, this estimation is incorrect. The passive perception misses the fine-grained visual details that waypoint 1 actually connects to a small doorway leading to a different room, while waypoint 2 is the correct corridor path. With proactive perception (top, green), the orchestration agent identifies missing information and generates a targeted visual query with focused regions. The perception agent examines waypoints 1 and 2 in detail. It discovers that waypoint 1 leads to a small doorway (side room) while waypoint 2 provides better corridor access. Value estimates are corrected: waypoint 1 drops to 0.2 and waypoint 2 rises to 0.9. This prevents a critical navigation error. This demonstrates that reasoning-guided queries and focused perception acquire instruction-relevant attributes missed by passive processing, enabling accurate semantic value estimation for reliable waypoint selection.

Figure~\ref{fig:qualitative_case} (right) demonstrates the advantage of BD-MCTS-guided top-$k$ focused reasoning in error correction. The navigation instruction requires sequential progression: approach the clock, turn right through a doorway, enter the doorway left of the bench, pass through the wooden entry, and stop at the table. The agent previously made an incorrect decision at nav\_3, leading to a bedroom area (nav\_11/nav\_14) that deviates from the correct path. Nav\_9 represents the central tiled pathway beyond nav\_7, serving as the primary route into the main hall.  With BD-MCTS (top, green), the decision agent receives top-$k$ candidates including nav\_9 and focuses reasoning on its associated historical context. By analyzing nav\_9's visual information describing it as "the primary route into the broader central space implying connection to areas with architectural features," the agent recognizes the path deviation and corrects the error by backtracking from nav\_13 to nav\_9. Without BD-MCTS (bottom, yellow), the agent lacks top-$k$ guidance and processes all historical candidates uniformly. At nav\_12, it observes a "wooden doorway" and directly matches it with the instruction's later-stage landmark "entry way made out of wood," ignoring that the current location (bedroom area) is irrelevant to the instruction stage (main hall area). This superficial visual matching causes the agent to continue on the incorrect path without reflecting on historical decisions. The case validates that top-$k$ focused reasoning enables global historical awareness and path correction, while undirected reasoning over all candidates leads to myopic reactive behavior that matches current observations to instruction fragments without considering path validity or navigation stage consistency.

\section{Conclusion}
\label{sec:conclusion}

This paper presents ProFocus, a training-free progressive framework for vision-and-language navigation. By unifying proactive perception and focused reasoning, ProFocus mitigates the issues of perception redundancy and unfocused reasoning in foundation model-based VLN. Experimental results on the R2R and REVERIE benchmarks demonstrate ProFocus's competitive zero-shot performance, with significant improvements in navigation success rates and path efficiency. 

The training-free design facilitates easy deployment in practical applications. Future work will extend ProFocus to long-horizon, multi-goal tasks in complex environments, such as robotic assistance for individuals with disabilities, where focused reasoning and proactive perception hold significant potential for enhancing efficiency and adaptability.
{
    \small
    \bibliographystyle{ieeenat_fullname}
    \bibliography{main}

\begin{thebibliography}{57}
\providecommand{\natexlab}[1]{#1}
\providecommand{\url}[1]{\texttt{#1}}
\expandafter\ifx\csname urlstyle\endcsname\relax
  \providecommand{\doi}[1]{doi: #1}\else
  \providecommand{\doi}{doi: \begingroup \urlstyle{rm}\Url}\fi

\bibitem[An et~al.(2025)An, Tian, Leng, Nie, Lin, Wang, Chen, Zhang, and Lu]{An2025AGLA}
Wenbin An, Feng Tian, Sicong Leng, Jiahao Nie, Haonan Lin, Qianying Wang, Ping Chen, Xiaoqin Zhang, and Shijian Lu.
\newblock Mitigating object hallucinations in large vision-language models with assembly of global and local attention.
\newblock In \emph{Proceedings of the IEEE/CVF Conference on Computer Vision and Pattern Recognition (CVPR)}, pages 29915--29926, 2025.

\bibitem[Anderson et~al.(2018)Anderson, Wu, Teney, Bruce, Johnson, S{\"u}nderhauf, Reid, Gould, and Van Den~Hengel]{anderson2018vision}
Peter Anderson, Qi Wu, Damien Teney, Jake Bruce, Mark Johnson, Niko S{\"u}nderhauf, Ian Reid, Stephen Gould, and Anton Van Den~Hengel.
\newblock Vision-and-language navigation: Interpreting visually-grounded navigation instructions in real environments.
\newblock In \emph{Proceedings of the IEEE/CVF Conference on Computer Vision and Pattern Recognition (CVPR)}, pages 3674--3683, 2018.

\bibitem[Bochkovskii et~al.(2025)Bochkovskii, Delaunoy, Germain, Santos, Zhou, Richter, and Koltun]{Bochkovskii2024:arxiv}
Aleksei Bochkovskii, Ama\"{e}l Delaunoy, Hugo Germain, Marcel Santos, Yichao Zhou, Stephan~R. Richter, and Vladlen Koltun.
\newblock Depth pro: Sharp monocular metric depth in less than a second.
\newblock In \emph{Proceedings of the International Conference on Learning Representations (ICLR)}, 2025.

\bibitem[Browne et~al.(2012)Browne, Powley, Whitehouse, Lucas, Cowling, Rohlfshagen, Tavener, Perez, Samothrakis, and Colton]{browne2012survey}
Cameron~B Browne, Edward Powley, Daniel Whitehouse, Simon~M Lucas, Peter~I Cowling, Philipp Rohlfshagen, Stephen Tavener, Diego Perez, Spyridon Samothrakis, and Simon Colton.
\newblock A survey of {Monte Carlo} tree search methods.
\newblock \emph{IEEE Transactions on Computational Intelligence and AI in Games}, 4\penalty0 (1):\penalty0 1--43, 2012.

\bibitem[Chen et~al.(2024)Chen, Lin, Xu, Chai, Liang, and Wong]{chen2024mapgpt}
Jiaqi Chen, Bingqian Lin, Ran Xu, Zhenhua Chai, Xiaodan Liang, and Kwan-Yee~K. Wong.
\newblock {MapGPT}: Map-guided prompting with adaptive path planning for vision-and-language navigation.
\newblock In \emph{Proceedings of the Annual Meeting of the Association for Computational Linguistics (ACL)}, 2024.

\bibitem[Chen et~al.(2025)Chen, Lin, Liu, Ma, Liang, and Wong]{chen2025affordances}
Jiaqi Chen, Bingqian Lin, Xinmin Liu, Lin Ma, Xiaodan Liang, and Kwan-Yee~K Wong.
\newblock Affordances-oriented planning using foundation models for continuous vision-language navigation.
\newblock In \emph{Proceedings of the AAAI Conference on Artificial Intelligence (AAAI)}, pages 23568--23576, 2025.

\bibitem[Chen et~al.(2021)Chen, Guhur, Schmid, and Laptev]{chen2021history}
Shizhe Chen, Pierre-Louis Guhur, Cordelia Schmid, and Ivan Laptev.
\newblock History aware multimodal transformer for vision-and-language navigation.
\newblock In \emph{Advances in Neural Information Processing Systems (NeurIPS)}, pages 5834--5847, 2021.

\bibitem[Chen et~al.(2022)Chen, Guhur, Tapaswi, Schmid, and Laptev]{chen2022think}
Shizhe Chen, Pierre-Louis Guhur, Makarand Tapaswi, Cordelia Schmid, and Ivan Laptev.
\newblock Think global, act local: Dual-scale graph transformer for vision-and-language navigation.
\newblock In \emph{Proceedings of the IEEE/CVF Conference on Computer Vision and Pattern Recognition (CVPR)}, pages 16537--16547, 2022.

\bibitem[Fried et~al.(2018)Fried, Hu, Cirik, Rohrbach, Andreas, Morency, Berg-Kirkpatrick, Saenko, Klein, and Darrell]{fried2018speaker}
Daniel Fried, Ronghang Hu, Volkan Cirik, Anna Rohrbach, Jacob Andreas, Louis-Philippe Morency, Taylor Berg-Kirkpatrick, Kate Saenko, Dan Klein, and Trevor Darrell.
\newblock Speaker-follower models for vision-and-language navigation.
\newblock In \emph{Advances in Neural Information Processing Systems (NeurIPS)}, 2018.

\bibitem[Hao et~al.(2020)Hao, Li, Li, Carin, and Gao]{hao2020towards}
Weituo Hao, Chunyuan Li, Xiujun Li, Lawrence Carin, and Jianfeng Gao.
\newblock Towards learning a generic agent for vision-and-language navigation via pre-training.
\newblock In \emph{Proceedings of the IEEE/CVF Conference on Computer Vision and Pattern Recognition (CVPR)}, pages 13137--13146, 2020.

\bibitem[Hong et~al.(2025)Hong, Yu, Gu, Wang, Gan, Tang, Cheng, Qi, Ji, Pan, et~al.]{hong2025glm}
Wenyi Hong, Wenmeng~English Yu, Xiaotao Gu, Guo Wang, Guobing Gan, Haomiao Tang, Jiale Cheng, Ji Qi, Junhui Ji, Lihang Pan, et~al.
\newblock {GLM-4.5V} and {GLM-4.1V-Thinking}: Towards versatile multimodal reasoning with scalable reinforcement learning.
\newblock \emph{arXiv preprint arXiv:2507.01006}, 2025.

\bibitem[Hong et~al.(2021)Hong, Wu, Qi, Rodriguez-Opazo, and Gould]{hong2021vln}
Yicong Hong, Qi Wu, Yuankai Qi, Cristian Rodriguez-Opazo, and Stephen Gould.
\newblock {VLN BERT}: A recurrent vision-and-language {BERT} for navigation.
\newblock In \emph{Proceedings of the IEEE/CVF Conference on Computer Vision and Pattern Recognition}, pages 1643--1653, 2021.

\bibitem[Huys et~al.(2012)Huys, Eshel, {O'Nions}, Sheridan, Dayan, and Roiser]{huys2012bonsai}
Quentin J.~M. Huys, Neir Eshel, Elizabeth {O'Nions}, Luke Sheridan, Peter Dayan, and Jonathan~P. Roiser.
\newblock Bonsai trees in your head: How the pavlovian system sculpts goal-directed choices by pruning decision trees.
\newblock \emph{PLOS Computational Biology}, 8\penalty0 (5):\penalty0 e1002410, 2012.

\bibitem[Lin et~al.(2025{\natexlab{a}})Lin, Nie, Loun~Zai, Wei, Han, Xu, Niu, Han, Lin, Lu, et~al.]{lin2025evolvenav}
Bingqian Lin, Yunshuang Nie, Khun Loun~Zai, Ziming Wei, Mingfei Han, Rongtao Xu, Minzhe Niu, Jianhua Han, Liang Lin, Cewu Lu, et~al.
\newblock {EvolveNav}: Self-improving embodied reasoning for {LLM}-based vision-language navigation.
\newblock \emph{arXiv preprint arXiv:2506.01551}, 2025{\natexlab{a}}.

\bibitem[Lin et~al.(2025{\natexlab{b}})Lin, Nie, Wei, Chen, Ma, Han, Xu, Chang, and Liang]{lin2025navcot}
Bingqian Lin, Yunshuang Nie, Ziming Wei, Jiaqi Chen, Shikui Ma, Jianhua Han, Hang Xu, Xiaojun Chang, and Xiaodan Liang.
\newblock {NavCoT}: Boosting {LLM}-based vision-and-language navigation via learning disentangled reasoning.
\newblock \emph{IEEE Transactions on Pattern Analysis and Machine Intelligence (TPAMI)}, 2025{\natexlab{b}}.

\bibitem[Lin et~al.(2025{\natexlab{c}})Lin, Long, Wan, Wang, Yang, Yang, Yao, Chen, Guo, Li, et~al.]{lin2025sail}
Lin Lin, Jiefeng Long, Zhihe Wan, Yuchi Wang, Dingkang Yang, Shuang Yang, Yueyang Yao, Xu Chen, Zirui Guo, Shengqiang Li, et~al.
\newblock Sail-embedding technical report: Omni-modal embedding foundation model.
\newblock \emph{arXiv preprint arXiv:2510.12709}, 2025{\natexlab{c}}.

\bibitem[Liu et~al.(2024{\natexlab{a}})Liu, Feng, Xue, Wang, Wu, Lu, et~al.]{deepseekai2024deepseekv3}
Aixin Liu, Bei Feng, Bing Xue, Bingxuan Wang, Bochao Wu, Chengda Lu, et~al.
\newblock Deepseek-v3 technical report.
\newblock \emph{arXiv preprint arXiv:2412.19437}, 2024{\natexlab{a}}.

\bibitem[Liu et~al.(2025{\natexlab{a}})Liu, Zhou, Zhang, Zhang, Huang, and Duan]{liu2025msnav}
Chenghao Liu, Zhimu Zhou, Jiachen Zhang, Minghao Zhang, Songfang Huang, and Huiling Duan.
\newblock {MSNav}: Zero-shot vision-and-language navigation with dynamic memory and {LLM} spatial reasoning.
\newblock \emph{arXiv preprint arXiv:2508.16654}, 2025{\natexlab{a}}.

\bibitem[Liu et~al.(2025{\natexlab{b}})Liu, Yang, Qian, Yin, Wang, Li, Liu, Zhai, Liu, and Zhang]{liu2025reinforcement}
Keliang Liu, Dingkang Yang, Ziyun Qian, Weijie Yin, Yuchi Wang, Hongsheng Li, Jun Liu, Peng Zhai, Yang Liu, and Lihua Zhang.
\newblock Reinforcement learning meets large language models: A survey of advancements and applications across the llm lifecycle.
\newblock \emph{arXiv preprint arXiv:2509.16679}, 2025{\natexlab{b}}.

\bibitem[Liu et~al.(2024{\natexlab{b}})Liu, Lin, Hewitt, Paranjape, Bevilacqua, Petroni, and Liang]{liu2024lostinthemiddle}
Nelson~F. Liu, Kevin Lin, John Hewitt, Ashwin Paranjape, Michele Bevilacqua, Fabio Petroni, and Percy Liang.
\newblock Lost in the middle: How language models use long contexts.
\newblock \emph{Transactions of the Association for Computational Linguistics}, 12:\penalty0 157--173, 2024{\natexlab{b}}.

\bibitem[Liu et~al.(2025{\natexlab{c}})Liu, Huang, Zhang, and Tang]{liu2025nav}
Qingxiang Liu, Ting Huang, Zeyu Zhang, and Hao Tang.
\newblock Nav-r1: Reasoning and navigation in embodied scenes.
\newblock \emph{arXiv preprint arXiv:2509.10884}, 2025{\natexlab{c}}.

\bibitem[Long et~al.(2024)Long, Li, Cai, and Dong]{long2024discuss}
Yuxing Long, Xiaoqi Li, Wenzhe Cai, and Hao Dong.
\newblock Discuss before moving: Visual language navigation via multi-expert discussions.
\newblock In \emph{IEEE International Conference on Robotics and Automation (ICRA)}, pages 17380--17387, 2024.

\bibitem[Mattar and Daw(2018)]{mattar2018prioritized}
Marcelo~G. Mattar and Nathaniel~D. Daw.
\newblock Prioritized memory access explains planning and hippocampal replay.
\newblock \emph{Nature Neuroscience}, 21\penalty0 (11):\penalty0 1609--1617, 2018.

\bibitem[Niu et~al.(2025)Niu, Ma, and Zhai]{niu2025research}
Wenying Niu, Hongru Ma, and Yue Zhai.
\newblock Research on task planning methods for embodied intelligence based on large language models.
\newblock In \emph{Proceedings of the IEEE International Symposium on Intelligent Robotics and Systems (ISoIRS)}, pages 1--5, 2025.

\bibitem[Qi et~al.(2020)Qi, Wu, Anderson, Wang, Wang, Shen, and van~den Hengel]{qi2020reverie}
Yuankai Qi, Qi Wu, Peter Anderson, Xin Wang, William~Yang Wang, Chunhua Shen, and Anton van~den Hengel.
\newblock {REVERIE}: Remote embodied visual referring expression in real indoor environments.
\newblock In \emph{Proceedings of the IEEE/CVF Conference on Computer Vision and Pattern Recognition (CVPR)}, pages 9982--9991, 2020.

\bibitem[Qiao et~al.(2025)Qiao, Lyu, Wang, Wang, Li, Zhang, Tan, and Wu]{qiao2025open}
Yanyuan Qiao, Wenqi Lyu, Hui Wang, Zixu Wang, Zerui Li, Yuan Zhang, Mingkui Tan, and Qi Wu.
\newblock Open-nav: Exploring zero-shot vision-and-language navigation in continuous environment with open-source llms.
\newblock In \emph{Proceedings of the IEEE International Conference on Robotics and Automation (ICRA)}, pages 6710--6717, 2025.

\bibitem[Qin et~al.(2025)Qin, Chen, Zhou, Chen, Li, Liao, Li, Che, and Yu]{qin2025survey}
Libo Qin, Qiguang Chen, Yuhang Zhou, Zhi Chen, Yinghui Li, Lizi Liao, Min Li, Wanxiang Che, and Philip~S Yu.
\newblock A survey of multilingual large language models.
\newblock \emph{Patterns}, 6\penalty0 (1), 2025.

\bibitem[Ravi et~al.(2024)Ravi, Gabeur, Hu, Hu, Ryali, Ma, Khedr, R{\"a}dle, Rolland, Gustafson, et~al.]{ravi2024sam}
Nikhila Ravi, Valentin Gabeur, Yuan-Ting Hu, Ronghang Hu, Chaitanya Ryali, Tengyu Ma, Haitham Khedr, Roman R{\"a}dle, Chloe Rolland, Laura Gustafson, et~al.
\newblock Sam 2: Segment anything in images and videos.
\newblock \emph{arXiv preprint arXiv:2408.00714}, 2024.

\bibitem[Song et~al.(2025)Song, Chen, Liu, Chen, Li, and Lin]{song2025towards}
Xinshuai Song, Weixing Chen, Yang Liu, Weikai Chen, Guanbin Li, and Liang Lin.
\newblock Towards long-horizon vision-language navigation: Platform, benchmark and method.
\newblock In \emph{Proceedings of the IEEE/CVF Conference on Computer Vision and Pattern Recognition (CVPR)}, pages 12078--12088, 2025.

\bibitem[Sutskever et~al.(2014)Sutskever, Vinyals, and Le]{sutskever2014sequence}
Ilya Sutskever, Oriol Vinyals, and Quoc~V Le.
\newblock Sequence to sequence learning with neural networks.
\newblock In \emph{Advances in Neural Information Processing Systems (NeurIPS)}, 2014.

\bibitem[Tan et~al.(2019)Tan, Yu, and Bansal]{tan2019learning}
Hao Tan, Licheng Yu, and Mohit Bansal.
\newblock Learning to navigate unseen environments: Back translation with environmental dropout.
\newblock In \emph{Proceedings of the Conference of the North American Chapter of the Association for Computational Linguistics: Human Language Technologies (NAACL-HLT)}, pages 2610--2621, 2019.

\bibitem[Vaswani et~al.(2017)Vaswani, Shazeer, Parmar, Uszkoreit, Jones, Gomez, Kaiser, and Polosukhin]{Vaswani2017}
Ashish Vaswani, Noam Shazeer, Niki Parmar, Jakob Uszkoreit, Llion Jones, Aidan~N. Gomez, {\L}ukasz Kaiser, and Illia Polosukhin.
\newblock Attention is all you need.
\newblock In \emph{Advances in Neural Information Processing Systems (NeurIPS)}, 2017.

\bibitem[Xu et~al.(2025{\natexlab{a}})Xu, Hao, Zong, Wang, Zhang, Wang, Lan, Gong, Ouyang, Meng, et~al.]{xu2025towards}
Fengli Xu, Qianyue Hao, Zefang Zong, Jingwei Wang, Yunke Zhang, Jingyi Wang, Xiaochong Lan, Jiahui Gong, Tianjian Ouyang, Fanjin Meng, et~al.
\newblock Towards large reasoning models: A survey of reinforced reasoning with large language models.
\newblock \emph{arXiv preprint arXiv:2501.09686}, 2025{\natexlab{a}}.

\bibitem[Xu et~al.(2025{\natexlab{b}})Xu, Gong, and Mu]{xu2025navq}
Peiran Xu, Xicheng Gong, and Yadong Mu.
\newblock {NavQ}: Learning a {Q}-model for foresighted vision-and-language navigation.
\newblock In \emph{Proceedings of the IEEE/CVF International Conference on Computer Vision (ICCV)}, pages 6327--6341, 2025{\natexlab{b}}.

\bibitem[Xue and He(2023)]{xue2023progressive}
Wei Xue and Hong He.
\newblock A progressive learning classifier based on dynamic differential weighted network for feature identification of brain network series.
\newblock \emph{Knowledge-Based Systems}, 274:\penalty0 110661, 2023.

\bibitem[Xue et~al.(2024)Xue, He, Wang, and Zhao]{xue2024sagn}
Wei Xue, Hong He, Yanbing Wang, and Ying Zhao.
\newblock Sagn: Sparse adaptive gated graph neural network with graph regularization for identifying dual-view brain networks.
\newblock \emph{IEEE Transactions on Neural Networks and Learning Systems}, 36\penalty0 (5):\penalty0 8085--8099, 2024.

\bibitem[Yang et~al.(2025{\natexlab{a}})Yang, Li, Yang, Zhang, Hui, Zheng, Yu, Gao, Huang, Lv, et~al.]{yang2025qwen3}
An Yang, Anfeng Li, Baosong Yang, Beichen Zhang, Binyuan Hui, Bo Zheng, Bowen Yu, Chang Gao, Chengen Huang, Chenxu Lv, et~al.
\newblock Qwen3 technical report.
\newblock \emph{arXiv preprint arXiv:2505.09388}, 2025{\natexlab{a}}.

\bibitem[Yang et~al.(2022{\natexlab{a}})Yang, Huang, Kuang, Du, and Zhang]{yang2022disentangled}
Dingkang Yang, Shuai Huang, Haopeng Kuang, Yangtao Du, and Lihua Zhang.
\newblock Disentangled representation learning for multimodal emotion recognition.
\newblock In \emph{Proceedings of the 30th ACM International Conference on Multimedia (ACM MM)}, pages 1642--1651, 2022{\natexlab{a}}.

\bibitem[Yang et~al.(2022{\natexlab{b}})Yang, Huang, Wang, Liu, Zhai, Su, Li, and Zhang]{yang2022emotion}
Dingkang Yang, Shuai Huang, Shunli Wang, Yang Liu, Peng Zhai, Liuzhen Su, Mingcheng Li, and Lihua Zhang.
\newblock Emotion recognition for multiple context awareness.
\newblock In \emph{Proceedings of the European Conference on Computer Vision (ECCV)}, pages 144--162, 2022{\natexlab{b}}.

\bibitem[Yang et~al.(2023)Yang, Chen, Wang, Wang, Li, Liu, Zhao, Huang, Dong, Zhai, and Zhang]{yang2023context}
Dingkang Yang, Zhaoyu Chen, Yuzheng Wang, Shunli Wang, Mingcheng Li, Siao Liu, Xiao Zhao, Shuai Huang, Zhiyan Dong, Peng Zhai, and Lihua Zhang.
\newblock Context de-confounded emotion recognition.
\newblock In \emph{Proceedings of the IEEE/CVF Conference on Computer Vision and Pattern Recognition (CVPR)}, pages 19005--19015, 2023.

\bibitem[Yang et~al.(2024{\natexlab{a}})Yang, Li, Qu, Yang, Zhai, Wang, and Zhang]{yang2024asynchronous}
Dingkang Yang, Mingcheng Li, Linhao Qu, Kun Yang, Peng Zhai, Song Wang, and Lihua Zhang.
\newblock Asynchronous multimodal video sequence fusion via learning modality-exclusive and-agnostic representations.
\newblock \emph{IEEE Transactions on Circuits and Systems for Video Technology}, 2024{\natexlab{a}}.

\bibitem[Yang et~al.(2024{\natexlab{b}})Yang, Li, Xiao, Liu, Yang, Chen, Wang, Zhai, Li, and Zhang]{yang2024MCIS}
Dingkang Yang, Mingcheng Li, Dongling Xiao, Yang Liu, Kun Yang, Zhaoyu Chen, Yuzheng Wang, Peng Zhai, Ke Li, and Lihua Zhang.
\newblock Towards multimodal sentiment analysis debiasing via bias purification.
\newblock In \emph{Proceedings of the European Conference on Computer Vision (ECCV)}, 2024{\natexlab{b}}.

\bibitem[Yang et~al.(2024{\natexlab{c}})Yang, Wei, Xiao, Wang, Wu, Li, Li, Wang, Chen, Jiang, et~al.]{yang2024pediatricsgpt}
Dingkang Yang, Jinjie Wei, Dongling Xiao, Shunli Wang, Tong Wu, Gang Li, Mingcheng Li, Shuaibing Wang, Jiawei Chen, Yue Jiang, et~al.
\newblock Pediatricsgpt: Large language models as chinese medical assistants for pediatric applications.
\newblock \emph{Advances in Neural Information Processing Systems (NeurIPS)}, 37:\penalty0 138632--138662, 2024{\natexlab{c}}.

\bibitem[Yang et~al.(2024{\natexlab{d}})Yang, Yang, Kuang, Chen, Wang, and Zhang]{yang2024towards}
Dingkang Yang, Kun Yang, Haopeng Kuang, Zhaoyu Chen, Yuzheng Wang, and Lihua Zhang.
\newblock Towards context-aware emotion recognition debiasing from a causal demystification perspective via de-confounded training.
\newblock \emph{IEEE Transactions on Pattern Analysis and Machine Intelligence}, 2024{\natexlab{d}}.

\bibitem[Yang et~al.(2024{\natexlab{e}})Yang, Yang, Li, Wang, Wang, and Zhang]{yang2024robust}
Dingkang Yang, Kun Yang, Mingcheng Li, Shunli Wang, Shuaibing Wang, and Lihua Zhang.
\newblock Robust emotion recognition in context debiasing.
\newblock In \emph{Proceedings of the IEEE/CVF Conference on Computer Vision and Pattern Recognition (CVPR)}, pages 12447--12457, 2024{\natexlab{e}}.

\bibitem[Yang et~al.(2025{\natexlab{b}})Yang, Li, Wu, Chen, Jiang, Liu, Zhai, and Zhang]{yang2025improvingmsa}
Dingkang Yang, Mingcheng Li, Xuecheng Wu, Zhaoyu Chen, Kaixun Jiang, Keliang Liu, Peng Zhai, and Lihua Zhang.
\newblock Improving multimodal sentiment analysis via modality optimization and dynamic primary modality selection.
\newblock \emph{arXiv preprint arXiv:2511.06328}, 2025{\natexlab{b}}.

\bibitem[Yang et~al.(2025{\natexlab{c}})Yang, Wei, Li, Liu, Liu, Hu, He, Ju, Zhou, Liu, et~al.]{yang2025medaide}
Dingkang Yang, Jinjie Wei, Mingcheng Li, Jiyao Liu, Lihao Liu, Ming Hu, Junjun He, Yakun Ju, Wei Zhou, Yang Liu, et~al.
\newblock Medaide: Information fusion and anatomy of medical intents via llm-based agent collaboration.
\newblock \emph{Information Fusion}, page 103743, 2025{\natexlab{c}}.

\bibitem[Yang et~al.(2025{\natexlab{d}})Yang, Xiao, Wei, Li, Chen, Li, and Zhang]{yang2025improving}
Dingkang Yang, Dongling Xiao, Jinjie Wei, Mingcheng Li, Zhaoyu Chen, Ke Li, and Lihua Zhang.
\newblock Improving factuality in large language models via decoding-time hallucinatory and truthful comparators.
\newblock In \emph{Proceedings of the AAAI Conference on Artificial Intelligence (AAAI)}, pages 25606--25614, 2025{\natexlab{d}}.

\bibitem[Yin et~al.(2025)Yin, Xu, Zhao, Wang, Zhou, and Lu]{yin2025unigoal}
Hang Yin, Xiuwei Xu, Linqing Zhao, Ziwei Wang, Jie Zhou, and Jiwen Lu.
\newblock {UniGoal}: Towards universal zero-shot goal-oriented navigation.
\newblock In \emph{Proceedings of the IEEE/CVF Conference on Computer Vision and Pattern Recognition (CVPR)}, pages 19057--19066, 2025.

\bibitem[Yu et~al.(2025)Yu, Long, Yang, Zeng, Fan, Zhang, and Dong]{yu2025correctnav}
Zhuoyuan Yu, Yuxing Long, Zihan Yang, Chengyan Zeng, Hongwei Fan, Jiyao Zhang, and Hao Dong.
\newblock Correctnav: Self-correction flywheel empowers vision-language-action navigation model.
\newblock \emph{arXiv preprint arXiv:2508.10416}, 2025.

\bibitem[Zhang et~al.(2025{\natexlab{a}})Zhang, Li, Qi, Li, Liu, Wang, Liu, Zhou, Wu, Li, et~al.]{zhang2025embodied}
Jiazhao Zhang, Anqi Li, Yunpeng Qi, Minghan Li, Jiahang Liu, Shaoan Wang, Haoran Liu, Gengze Zhou, Yuze Wu, Xingxing Li, et~al.
\newblock Embodied navigation foundation model.
\newblock \emph{arXiv preprint arXiv:2509.12129}, 2025{\natexlab{a}}.

\bibitem[Zhang et~al.(2025{\natexlab{b}})Zhang, Wang, Wang, Li, Liu, Wei, Wang, Zhang, and Wang]{zhang2024uni}
Jiazhao Zhang, Kunyu Wang, Shaoan Wang, Minghan Li, Haoran Liu, Songlin Wei, Zhongyuan~champion Wang, Zhizheng Zhang, and He Wang.
\newblock {Uni-NaVid}: A video-based vision-language-action model for unifying embodied navigation tasks.
\newblock In \emph{Proceedings of the Robotics: Science and Systems (RSS)}, 2025{\natexlab{b}}.

\bibitem[Zhang et~al.(2025{\natexlab{c}})Zhang, Su, Wu, An, Zhang, Wang, Wang, Ding, Zhao, and Li]{zhang2025cross}
Pingrui Zhang, Yifei Su, Pengyuan Wu, Dong An, Li Zhang, Zhigang Wang, Dong Wang, Yan Ding, Bin Zhao, and Xuelong Li.
\newblock Cross from left to right brain: Adaptive text dreamer for vision-and-language navigation.
\newblock \emph{arXiv preprint arXiv:2505.20897}, 2025{\natexlab{c}}.

\bibitem[Zhang et~al.(2024)Zhang, Ma, Li, Qiao, Wang, Chai, Wu, Bansal, and Kordjamshidi]{zhang2024visionandlanguage}
Yue Zhang, Ziqiao Ma, Jialu Li, Yanyuan Qiao, Zun Wang, Joyce Chai, Qi Wu, Mohit Bansal, and Parisa Kordjamshidi.
\newblock Vision-and-language navigation today and tomorrow: A survey in the era of foundation models.
\newblock \emph{Transactions on Machine Learning Research (TMLR)}, 2024.

\bibitem[Zheng et~al.(2024)Zheng, Huang, Zhao, Zhong, and Wang]{10657191}
Duo Zheng, Shijia Huang, Lin Zhao, Yiwu Zhong, and Liwei Wang.
\newblock Towards learning a generalist model for embodied navigation.
\newblock In \emph{2024 IEEE/CVF Conference on Computer Vision and Pattern Recognition (CVPR)}, pages 13624--13634, 2024.

\bibitem[Zhou et~al.(2024)Zhou, Hong, and Wu]{zhou2024navgpt}
Gengze Zhou, Yicong Hong, and Qi Wu.
\newblock {NavGPT}: Explicit reasoning in vision-and-language navigation with large language models.
\newblock In \emph{Proceedings of the AAAI Conference on Artificial Intelligence (AAAI)}, pages 7641--7649, 2024.

\bibitem[Zhu et~al.(2020)Zhu, Zhu, Chang, and Liang]{zhu2020vision}
Fengda Zhu, Yi Zhu, Xiaojun Chang, and Xiaodan Liang.
\newblock Vision-language navigation with self-supervised auxiliary reasoning tasks.
\newblock In \emph{Proceedings of the IEEE/CVF Conference on Computer Vision and Pattern Recognition (CVPR)}, pages 10012--10022, 2020.

\end{thebibliography}
}

\end{document}